# Fingerprint Liveness Detection Based on Quality Measures

Javier Galbally, Fernando Alonso-Fernandez, Julian Fierrez, and Javier Ortega-Garcia
Biometric Recognition Group–ATVS, EPS, Universidad Autonoma de Madrid,
C/ Francisco Tomas y Valiente 11, 28049 Madrid, Spain
Email: {javier.galbally, fernando.alonso, julian.fierrez, javier.ortega}@uam.es

*Abstract*—A new fingerprint parameterization for liveness detection based on quality measures is presented. The novel feature set is used in a complete liveness detection system and tested on the development set of the LivDET competition, comprising over 4,500 real and fake images acquired with three different optical sensors. The proposed solution proves to be robust to the multi-sensor scenario, and presents an overall rate of 93% of correctly classified samples. Furthermore, the liveness detection method presented has the added advantage over previously studied techniques of needing just one image from a finger to decide whether it is real or fake.

## I. INTRODUCTION

In the last recent years important research efforts have been conducted to study the vulnerabilities of biometric systems to direct attacks to the sensor (carried out using synthetic biometric traits such as gummy fingers or high quality iris printed images) [1], [2], and indirect attacks (carried out against some of the inner modules of the system) [3]. Furthermore, the interest for the analysis of security vulnerabilities has surpassed the scientific field and different standardization initiatives at international level have emerged in order to deal with the problem of security evaluation in biometric systems, such as the Common Criteria through different Supporting Documents [4], or the Biometric Evaluation Methodology [5].

Within the studied vulnerabilities, special attention has been paid to direct attacks carried out against fingerprint recognition systems [6], [7], [8]. These attacking methods consist on presenting a synthetically generated fingerprint to the sensor so that it is recognized as the legitimate user and access is granted. These attacks have the advantage over other more sophisticated attacking algorithms, such as the hill-climbing strategies [9], of not needing any information about the internal working of the system (e.g., features used, template format). Furthermore, as they are carried out outside the digital domain these attacks are more difficult to be detected as the digital protection mechanisms (e.g., digital signature, watermarking) are not valid to prevent them.

These research efforts in the study of the vulnerabilities of automatic recognition systems to direct attacks have led to an enhancement of the security level offered by biometric systems through the proposal of specific countermeasures. In particular, different liveness detection methods have been presented through the past recent years. These algorithms are anti-spoofing techniques which use different physiological properties to distinguish between real and fake traits, thus improving the robustness of the system against direct attacks. Fingerprint liveness detection approaches can broadly be divided into:

- Software-based techniques. In this case fake fingerprints are detected once the sample has been acquired with a standard sensor (i.e., features used to distinguish between real and fake fingers are extracted from the fingerprint image, and not from the finger itself). These approaches include the use of skin perspiration [10], or skin elasticity properties [11], [12].
- Hardware-based techniques. In this case some specific device is added to the sensor in order to detect particular properties of a living finger such as the blood pressure [13], the odor [14], or the heartbeat [15].

Software-based techniques have the advantage over the hardware-based ones of being less expensive (as no extra device in needed), and less intrusive for the user (very important characteristic for a practical liveness detection solution).

A comparative analysis of different software-based solutions for liveness detection is presented in [16]. The authors study the efficiency of several approaches and give an estimation of the best performing static and dynamic features for vitality detection. Static features are those which require two or more fingerprint impressions (i.e., the finger is placed and lifted from the sensor several times) of the same finger, while dynamic features are extracted from multiple image frames (i.e., the finger is placed on the sensor for a sort time and different images are acquired).

In the present work we propose a new parameterization based on quality measures for a software-based solution in fingerprint liveness detection. This novel strategy has the clear advantage over the previously proposed methods of needing just one fingerprint image (i.e., the same fingerprint image used for access) to extract the necessary features in order to determine if the finger presented to the sensor is real or fake. This fact shortens the acquisition process and reduces the inconvenience for the final user. The presented method has been tested on the database provided as development set in the Fingerprint Liveness Detection Competition LivDET 2009 [17], comprising over 4,500 real and fake samples generated with different materials and captured with different sensors. The liveness detection system presented, using the proposed parameterization reached a significant 93.4% of correctly

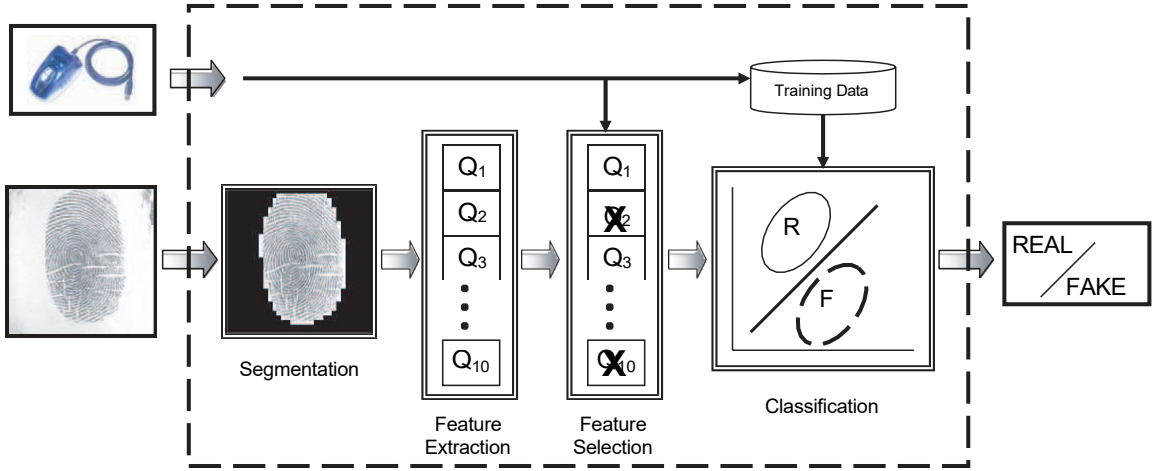

Fig. 1. General diagram of the liveness detection system presented in this work.

classified (real or fake) samples.

The rest of the paper is structured as follows. The liveness detection system is described in Sect. II, with special attention to the novel parameterization proposed. In Sect. III the database used in the experiments is presented, and results are given in Sect. IV. Conclusions are finally drawn in Sect. V.

## II. LIVENESS DETECTION SYSTEM

The problem of liveness detection can be seen as a two-class classification problem where an input fingerprint image has to be assigned to one of two classes: real or fake. The key point of the process is to find a set of discriminant features which permits to build an appropriate classifier which gives the probability of the image vitality given the extracted set of features. In the present work we propose a novel parameterization using quality measures which is tested on a complete liveness detection system.

A general diagram of the liveness detection system presented in this work is shown in Fig. 1. Two inputs are given to the system: $i$) the fingerprint image to be classified, and $ii$) the sensor used in the acquisition process.

In the first step the fingerprint is segmented from the background, for this purpose, Gabor filters are used as proposed in [18]. Once the useful information of the total image has been separated, ten different quality measures are extracted which will serve as the feature vector that will be used in the classification. Prior to the classification step, the best performing features are selected depending on the sensor that was used in the acquisition. Once the final feature vector has been generated the fingerprint is classified as real (generated by a living finger), or fake (coming from a gummy finger), using as training data of the classifier the dataset corresponding to the acquisition sensor.

### A. Feature Extraction

The parameterization proposed in the present work and applied to liveness detection comprises ten quality-based features. A number of approaches for fingerprint image quality

**Fingerprint Image Quality Estimation Methods**

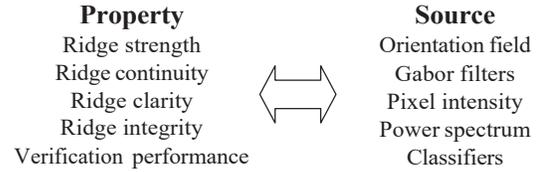

| Property | Source |
|---|---|
| Ridge strength | Orientation field |
| Ridge continuity | Gabor filters |
| Ridge clarity | Pixel intensity |
| Ridge integrity | Power spectrum |
| Verification performance | Classifiers |

Fig. 2. Taxonomy of the different approaches for fingerprint image quality computation that have been described in the literature.

computation have been described in the literature. A taxonomy is given in [19] (see Fig. 2). Image quality can be assessed by measuring one of the following properties: ridge strength or directionality, ridge continuity, ridge clarity, integrity of the ridge-valley structure, or estimated verification performance when using the image at hand. A number of sources of information are used to measure these properties: $i$) angle information provided by the direction field, $ii$) Gabor filters, which represent another implementation of the direction angle [20], $iii$) pixel intensity of the gray-scale image, $iv$) power spectrum, and $v$) Neural Networks. Fingerprint quality can be assessed either analyzing the image in a holistic manner, or combining the quality from local non-overlapped blocks of the image.

In the following, we give some details about the quality measures used in this paper. We have implemented several measures that make use of the above mentioned properties for quality assessment, see Table I:

*1) Ridge-strength measures:*

- **Orientation Certainty Level** ($Q_{OCL}$) [21], which measures the energy concentration along the dominant direction of ridges using the intensity gradient. It is computed as the ratio between the two eigenvalues of the covariance matrix of the gradient vector. A relative weight is given to each region of the image based on its distance from the centroid, since regions near the centroid are supposed to provide more reliable information [22]. An example of Orientation Certainty Level computation is shown in Fig. 3 for two fingerprints of different quality.

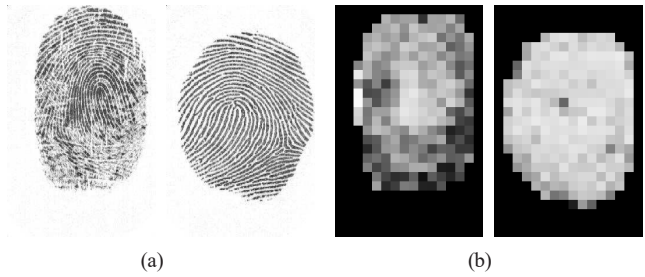
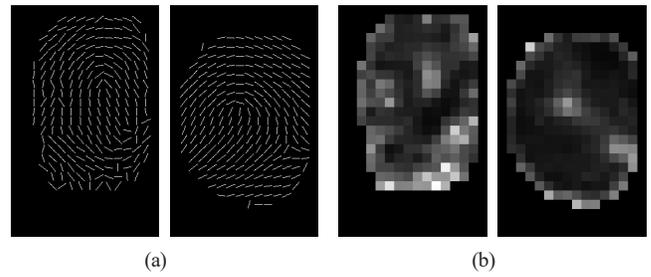

Fig. 3. Computation of the Orientation Certainty Level (*OCL*) for two fingerprints of different quality. Panel (a) are the input fingerprint images. Panel (b) are the block-wise values of the *OCL*; blocks with brighter color indicate higher quality in the region.

Fig. 5. Computation of the Local Orientation Quality (*LOQ*) for two fingerprints of different quality. Panel (a) are the direction fields of the images shown in Figure 3a. Panel (b) are the block-wise values of the average absolute difference of local orientation with the surrounding blocks; blocks with brighter color indicate higher difference value and thus, lower quality.

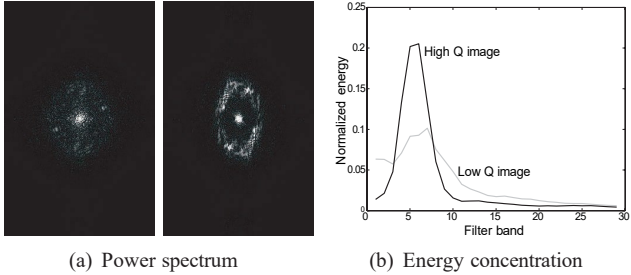

(a) Power spectrum     (b) Energy concentration

Fig. 4. Computation of the energy concentration in the power spectrum for two fingerprints of different quality. Panel (a) are the power spectra of the images shown in Figure 3. Panel (b) shows the energy distributions in the region of interest. The quality values for the low and high quality image are 0.35 and 0.88 respectively.

- **Energy concentration in the power spectrum ($Q_{ENERGY}$)** [22], which is computed using ring-shaped bands. For this purpose, a set of bandpass filters is employed to extract the energy in each frequency band. High quality images will have the energy concentrated in few bands while poor ones will have a more diffused distribution. The energy concentration is measured using the entropy. An example of quality estimation using the global quality index $Q_{ENERGY}$ is shown in Fig. 4 for two fingerprints of different quality.

*2) Ridge-continuity measures:*

- **Local Orientation Quality ($Q_{LOQ}$)** [23], which is computed as the average absolute difference of direction angle with the surrounding image blocks, providing information about how smoothly direction angle changes from block to block. Quality of the whole image is finally computed by averaging all the Local Orientation Quality scores of the image. In high quality images, it is expected that ridge direction changes smoothly across the whole image. An example of Local Orientation Quality computation is shown in Fig. 5 for two fingerprints of different quality.
- **Continuity of the orientation field ($Q_{COF}$)** [21]. This method relies on the fact that, in good quality images, ridges and valleys must flow sharply and smoothly in a locally constant direction. The direction change along rows and columns of the image is examined. Abrupt direction changes between consecutive blocks are then accumulated and mapped into a quality score. As we can observe in Fig. 5, ridge direction changes smoothly across the whole image in case of high quality.

*3) Ridge-clarity measures:*

- **Mean ($Q_{MEAN}$)** and **standard deviation ($Q_{ST\ D}$)** values of the gray level image, computed from the segmented foreground only. These two features had already been considered for liveness detection in [16].
- **Local Clarity Score ($Q_{LCS1}$ and $Q_{LCS2}$)** [23]. The sinusoidal-shaped wave that models ridges and valleys [24] is used to segment ridge and valley regions (see Figure 6). The clarity is then defined as the overlapping area of the gray level distributions of segmented ridges and valleys. For ridges/valleys with high clarity, both distributions should have a very small overlapping area. An example of quality estimation using the Local Clarity Score is shown in Fig. 7 for two fingerprint blocks of different quality. It should be noted that sometimes the sinusoidal-shaped wave cannot be extracted reliably, specially in bad quality regions of the image. The quality measure $Q_{LCS1}$ discards these regions, therefore being an optimistic measure of quality. This is compensated with $Q_{LCS2}$, which does not discard these regions, but they are assigned the lowest quality level.
- **Amplitude and variance of the sinusoid that models ridges and valleys ($Q_A$ and $Q_{V\ AR}$)** [24]. Based on these parameters, blocks are classified as *good* and *bad*. The quality of the fingerprint is then computed as the percentage of foreground blocks marked as *good*.

*B. Feature Selection*

Due to the curse of dimensionality, it is possible that the best classifying results are not obtained using the set of ten proposed features, but a subset of them. As we are dealing with a ten dimensional problem there are $2^{10} - 1 = 1,023$ possible feature subsets, which is a reasonably low number to apply exhaustive search as feature selection technique in order to find the best performing feature subset. This way we guarantee that we find the optimal set of features out of all the possible ones. The feature selection depends on the acquisition device (as shown in Fig. 1), as the optimal feature subsets might be different for different sensors.

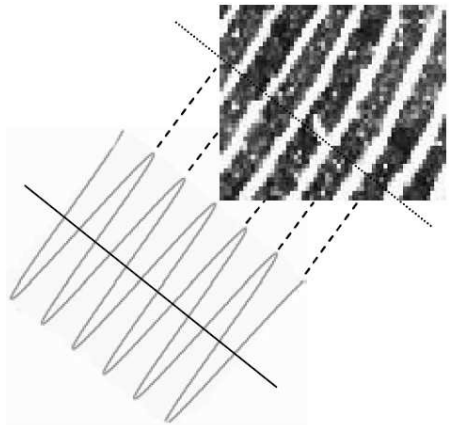

Fig. 6. Modeling of ridges and valleys as a sinusoid.

| Quality measure | Property measured | Source |
|---|---|---|
| $Q_{OCL}$ | Ridge strength | Local angle |
| $Q_E$ | Ridge strength | Power spectrum |
| $Q_{LOQ}$ | Ridge continuity | Local angle |
| $Q_{COF}$ | Ridge continuity | Local angle |
| $Q_{MEAN}$ | Ridge clarity | Pixel intensity |
| $Q_{STD}$ | Ridge clarity | Pixel intensity |
| $Q_{LCS1}$ | Ridge clarity | Pixel intensity |
| $Q_{LCS2}$ | Ridge clarity | Pixel intensity |
| $Q_A$ | Ridge clarity | Pixel intensity |
| $Q_{VAR}$ | Ridge clarity | Pixel intensity |

TABLE I
SUMMARY OF THE QUALITY MEASURES USED IN THE PARAMETERIZATION APPLIED TO FINGERPRINT LIVENESS DETECTION.

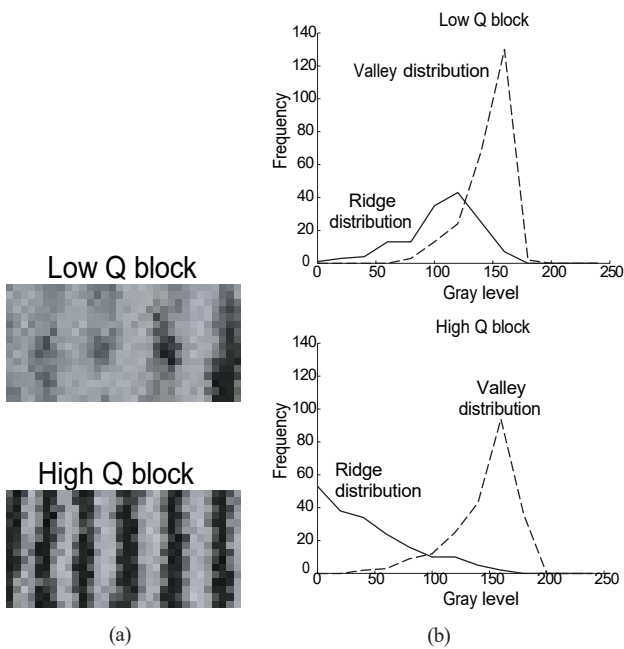

Fig. 7. Computation of the Local Clarity Score for two fingerprint blocks of different quality. Panel (a) shows the fingerprint blocks. Panel (b) shows the gray level distributions of the segmented ridges and valleys. The degree of overlapping for the low and high quality block is 0.22 and 0.10, respectively.

### C. Classifier

We have used Linear Discriminant Analysis (LDA) as classifier [25]. In the experiments the leave-one-out technique has been used, where all the samples acquired with the same sensor, except the one being classified, are used to fit the two normal distributions representing each of the classes. The sample being classified (which was left out of the training process) is then assigned to the most probable class.

### III. DATABASE

The database used in the experiments is the development set provided in the Fingerprint Liveness Detection Competition, LivDET 2009 [17]. It comprises three datasets of real and fake fingerprints (generated with different materials) captured each of them with a different optical sensor:

- Biometrika FX2000 (569 dpi). This dataset comprises 520 real and 520 fake images. The latter were generated with gummy fingers made of silicone.
- CrossMatch Verifier 300CL (500 dpi). This dataset comprises 1,000 real and 1,000 fake images. The latter were generated with gummy fingers made of silicone (310), gelatin (344), and playdoh (346).
- Identix DFR2100 (686 dpi). This dataset comprises 750 real and 750 fake images. The fake images were generated with gummy fingers made of silicone (250), gelatin (250), and playdoh (250).

The material with which the different fake images are made is known, however this fact is not used in anyway by the liveness detection system as in a real case this information would not be available to the application. Thus, as will be explained in the experiments, the feature selection is just made in terms of the sensor used in the acquisition.

In Fig. 8 we show some typical examples of the real and fake fingerprint images that can be found in the database (not necessarily belonging to the same subject). The fake fingerprints corresponding to the CrossMatch and Identix datasets were generated with each of the different materials. It can be noticed from the examples shown in Fig. 8 the difficulty of the classification problem, as even for a human expert would not be easy to distinguish between the real and fake samples present at the database.

### IV. RESULTS

The first objective of the experiments is to find the optimal feature subsets (out of the proposed 10 feature set) for each of the three datasets comprised in the database. Then the classification performance of each of the optimal subsets is computed on each of the datasets in terms of the Average Classification Error which is defined as $ACE = (FAR + FRR)/2$, where the FAR (False Acceptance Rate) represents the percentage

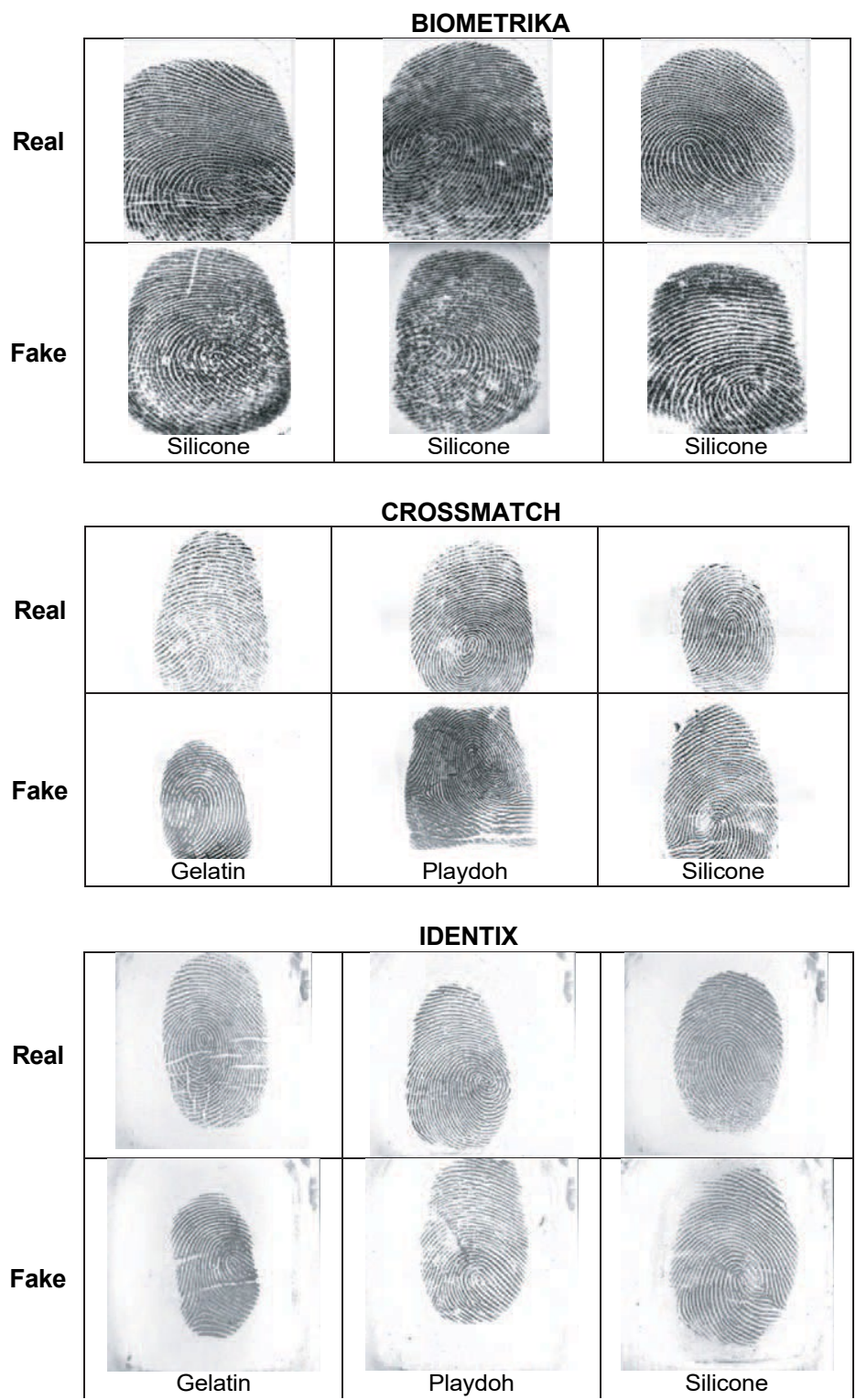

Fig. 8. Typical examples of real and fake fingerprint images that can be found in the database used in the experiments.

| Best feature subsets for Biometrika | | | | | | | | | | |
|---|---|---|---|---|---|---|---|---|---|---|
| | *Ridge Strength* | | *Ridge Continuity* | | *Ridge Clarity* | | | | | |
| # features | **$Q_{OCL}$** | **$Q_E$** | **$Q_{LOQ}$** | $Q_{COF}$ | **$Q_{MEAN}$** | $Q_{STD}$ | $Q_{LCS1}$ | $Q_{LCS2}$ | $Q_A$ | $Q_{VAR}$ | ACE (%) |
| 1 | | 1 | | | | | | | | | 21.83 |
| 2 | | 1 | | | | | | 1 | | | 13.37 |
| 3 | | 1 | | | 1 | | | 1 | | | 7.60 |
| 4 | 1 | 1 | 1 | | 1 | | | | | | 4.71 |
| 5 | 1 | 1 | 1 | | | 1 | | | 1 | | 2.60 |
| 6 | 1 | 1 | 1 | | 1 | 1 | | | 1 | | 2.12 |
| 7 | 1 | **1** | 1 | 1 | 1 | 1 | | | | 1 | 1.73 |
| 8 | 1 | 1 | 1 | 1 | 1 | 1 | | 1 | | 1 | 1.83 |
| 9 | 1 | 1 | 1 | 1 | 1 | 1 | | 1 | 1 | 1 | 2.02 |
| 10 | 1 | 1 | 1 | 1 | 1 | 1 | 1 | 1 | 1 | 1 | 2.31 |

| Best feature subsets for CrossMatch | | | | | | | | | | |
|---|---|---|---|---|---|---|---|---|---|---|
| | *Ridge Strength* | | *Ridge Continuity* | | *Ridge Clarity* | | | | | |
| # features | $Q_{OCL}$ | **$Q_E$** | $Q_{LOQ}$ | $Q_{COF}$ | **$Q_{MEAN}$** | $Q_{STD}$ | $Q_{LCS1}$ | **$Q_{LCS2}$** | $Q_A$ | $Q_{VAR}$ | ACE (%) |
| 1 | | | | | 1 | | | | | | 17.65 |
| 2 | | | | | 1 | | | | | 1 | 13.25 |
| 3 | | | | | 1 | | 1 | | 1 | 1 | 11.80 |
| 4 | | 1 | | | 1 | | | 1 | | 1 | 11.30 |
| 5 | | 1 | | | 1 | 1 | | 1 | | 1 | 11.45 |
| 6 | 1 | **1** | | | 1 | 1 | 1 | 1 | | | 11.15 |
| 7 | 1 | 1 | 1 | | 1 | 1 | 1 | 1 | | | 11.35 |
| 8 | 1 | 1 | | 1 | 1 | 1 | 1 | 1 | 1 | | 11.55 |
| 9 | 1 | 1 | 1 | 1 | 1 | 1 | 1 | 1 | | 1 | 11.95 |
| 10 | 1 | 1 | 1 | 1 | 1 | 1 | 1 | 1 | 1 | 1 | 12.80 |

| Best feature subsets for Identix | | | | | | | | | | |
|---|---|---|---|---|---|---|---|---|---|---|
| | *Ridge Strength* | | *Ridge Continuity* | | *Ridge Clarity* | | | | | |
| # features | $Q_{OCL}$ | $Q_E$ | $Q_{LOQ}$ | $Q_{COF}$ | $Q_{MEAN}$ | **$Q_{STD}$** | **$Q_{LCS1}$** | **$Q_{LCS2}$** | $Q_A$ | $Q_{VAR}$ | ACE (%) |
| 1 | | | | | 1 | | | | | | 20.07 |
| 2 | | 1 | | | | 1 | | | | | 11.93 |
| 3 | | | | | | 1 | 1 | 1 | | | 9.40 |
| 4 | | | | 1 | | 1 | 1 | 1 | | | 7.67 |
| 5 | 1 | | 1 | | | 1 | 1 | 1 | | | 7.20 |
| 6 | | | | 1 | | 1 | 1 | 1 | 1 | 1 | 7.07 |
| 7 | | **1** | 1 | | | 1 | 1 | 1 | 1 | 1 | 6.87 |
| 8 | | 1 | | 1 | 1 | 1 | 1 | 1 | 1 | 1 | 6.93 |
| 9 | 1 | 1 | 1 | 1 | | 1 | 1 | 1 | 1 | 1 | 7.13 |
| 10 | 1 | 1 | 1 | 1 | 1 | 1 | 1 | 1 | 1 | 1 | 7.20 |

TABLE II

BEST PERFORMING SUBSETS WITH AN INCREASING NUMBER OF FEATURES. A 1 MEANS THAT THE FEATURE IS INCLUDED, AND A BLANK SPACE THAT IT IS DISCARDED. THE OPTIMAL FEATURE SUBSET FOR EACH OF THE DATASETS IS HIGHLIGHTED IN GREY. THE BEST PERFORMING FEATURES ARE PRESENTED IN **BOLD**.

of fake fingerprints misclassified as real, and the FRR (False Rejection Rate) computes the percentage of real fingerprints assigned to the fake class.

*A. Feature Selection Results*

In order to find the optimal feature subsets, for each of the three datasets in the database, the classification performance of each of the 1,023 possible feature subsets was computed using the leave-one-out technique (i.e., all the samples in the dataset are used to train the classifier except the one being classified). The best feature subsets (for an increasing number of features) found for each of the sensors are shown in Table II, where a 1 means that the feature is included in the subset. The Average Classification Error for each of the best subsets is shown on the right, and the optimal feature subset is highlighted in grey.

From the results shown in Table II we can see that the most discriminant features for the Biometrika dataset are those measuring the ridge strength. Also, one ridge continuity ($Q_{LOQ}$) and one ridge clarity ($Q_{MEAN}$) measure are shown to provide certain discriminative capabilities with this sensor. In the case of the CrossMatch sensor, on the other hand, the least useful features for liveness detection are the ridge continuity related, while the ridge strength and ridge clarity measures have a similar importance (only $Q_{MEAN}$ clearly stands out). In the Identix dataset we can see that the best features are the ridge clarity related (specially $Q_{STD}$, $Q_{LCS1}$, and $Q_{LCS2}$), and, on the other hand, the ridge strength related are the least discriminant. The information extracted from Table II on the discriminant capabilities of the different parameters according to the ridge property measured is summarized in Table III.

The evolution of the ACE produced by each of the best feature subsets (right column in Table II) and for the three datasets is shown in Fig. 9, where the optimal error for each dataset is highlighted with a horizontal dashed line. In Fig. 9 we can see that the proposed parameterization is specially effective for liveness detection with the Biometrika sensor

|              | Ridge Strength      | Ridge Continuity | Ridge Clarity                          |
| ------------ | ------------------- | ---------------- | -------------------------------------- |
| Biometrika   | High ($Q_E$, $Q_{OCL}$) | Medium ($Q_{LOQ}$) | Medium ($Q_{MEAN}$)                    |
| CrossMatch   | Medium ($Q_E$)      | Low              | High ($Q_{MEAN}$, $Q_{LCS2}$)          |
| Identix      | Low                 | Medium           | High ($Q_{STD}$, $Q_{LCS1}$, $Q_{LCS2}$) |

TABLE III
SUMMARY FOR THE THREE DATASETS OF THE PARAMETERS DISCRIMINANT POWER ACCORDING TO THE RIDGE PROPERTY MEASURED. THE BEST PERFORMING FEATURES ARE SPECIFIED IN EACH CASE.

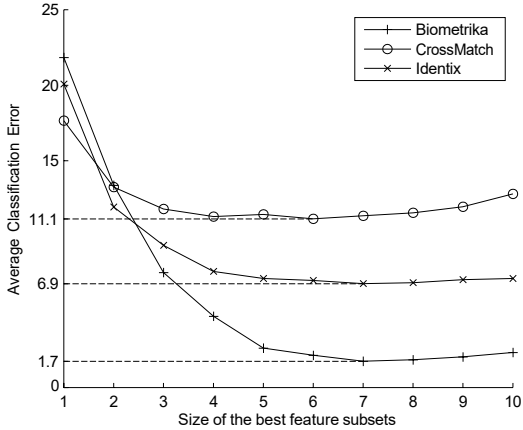

Fig. 9. Evolution of the ACE for the best feature subsets with an increasing number of features, and for the three datasets.

|            | Best subset for Biometrika $Q_{OCL}$, $Q_E$, $Q_{LOQ}$, $Q_{COF}$, $Q_{MEAN}$, $Q_{STD}$, $Q_{VAR}$ | | |
| ---------- | ------- | ------- | ------- |
|            | FAR (%) | FRR (%) | ACE (%) |
| Biometrika | 2.12    | 1.54    | 1.83    |
| CrossMatch | 12.48   | 12.32   | 12.40   |
| Identix    | 6.40    | 10.67   | 8.53    |
| TOTAL      | 7.00    | 8.17    | 7.58    |

(a) Performance of the best feature subset for the Biometrika dataset.

|            | Best subset for CrossMatch $Q_{OCL}$, $Q_E$, $Q_{MEAN}$, $Q_{STD}$, $Q_{LCS1}$, $Q_{LCS2}$ | | |
| ---------- | ------- | ------- | ------- |
|            | FAR (%) | FRR (%) | ACE (%) |
| Biometrika | 6.73    | 2.50    | 4.62    |
| CrossMatch | 10.30   | 11.94   | 11.12   |
| Identix    | 6.27    | 11.47   | 8.87    |
| TOTAL      | 7.76    | 8.63    | 8.12    |

(b) Performance of the best feature subset for the CrossMatch dataset.

|            | Best subset for Identix $Q_E$, $Q_{LOQ}$, $Q_{STD}$, $Q_{LCS1}$, $Q_{LCS2}$, $Q_A$, $Q_{VAR}$ | | |
| ---------- | ------- | ------- | ------- |
|            | FAR (%) | FRR (%) | ACE (%) |
| Biometrika | 6.92    | 0.96    | 3.94    |
| CrossMatch | 11.42   | 11.98   | 11.70   |
| Identix    | 6.40    | 7.07    | 6.73    |
| TOTAL      | 8.24    | 6.67    | 7.45    |

(c) Performance of the best feature subset for the Identix dataset.

TABLE IV
PERFORMANCE IN TERMS OF THE AVERAGE CLASSIFICATION ERROR (ACE) OF EACH OPTIMAL FEATURE SUBSET FOR THE BIOMETRIKA (a), CROSSMATCH (b), AND IDENTIX (c) DATASETS. THE BEST ACE FOR THE DIFFERENT DATASETS IS HIGHLIGHTED IN GREY.

where the ACE rapidly decreases when new features are added, while for the other two sensors the improvement in the error classification rate is smaller (in particular in the case of the CrossMatch).

B. Optimal Feature Subsets

Considering only the optimal feature subsets found for each of the sensors (highlighted in grey in Table II), we can see that the two most consistent features (that are included in the best subset for all the datasets) are $Q_E$ and $Q_{STD}$ (highlighted with a bold **1** in Table II). On the other hand, there is no feature that is not included at least in one of the optimal subsets which indicates that all the proposed features are relevant for fingerprint liveness detection.

The classification performance of each of the optimal feature subsets was computed for the three datasets, again using the leave-one-out technique. Results for each of the subsets are given in Table IV where the best result (the one corresponding to the optimal subset of a certain dataset, used to classify the images in that same dataset) are highlighted in grey.

From the results shown in Table IV we can see that the optimal combination of features that generalized best to all the sensors is the one corresponding to the Identix dataset as it produces the lowest total ACE (7.45%). However, all the optimal feature subsets have proven to be robust in the three datasets as the total ACE does not differ greatly.

The results also show that the new parameterization proposed performs best on the dataset captured with the Biometrika sensor where, for the optimal feature subset, an ACE of 1.73% is reached (over 98% of correctly classified samples). This result clearly improves the one presented in [16] where, on a very similar dataset and using a parameterization based on different static and dynamic features (which need several images to be extracted), a best 17% classification error is reported (almost 10 times higher than the error rate reached with our proposed quality-based approach).

On the other hand, the worst classification rate of our system is always generated on the CrossMatch dataset with a 11.12% of misclassified samples in the best case. An intermediate performance between the Biometrika and the CrossMatch datasets is reached for the Identix dataset in all cases.

Assuming that we can use for each of the datasets their own optimal feature subset (which is not a strong constraint as we should know the sensor used by the system), then the total ACE would be the average of the cells highlighted in grey in Table IV, and the system would present an optimal **ACE=6.56%**. This means that the system described in this work, using the new parameterization proposed, can correctly classify 93.44% of the fingerprint images available in the database, using just one single sample.

## V. CONCLUSIONS

A novel fingerprint parameterization for liveness detection based on quality measures has been proposed. The feature set has been used in a complete liveness detection system, and tested on the development set of the recent LivDET competition [17]. This challenging database comprises over 4,500 real and fake fingerprint images (generated with different synthetic materials), acquired with three optical sensors. The novel approach has proven to be robust to the multi-sensor scenario, correctly classifying (real or fake) over 93% of the fingerprint images.

The proposed approach is part of the software-based solutions as it distinguishes between images produced by real and fake fingers based only on the acquired sample, and not on other physiological measures (e.g., odor, heartbeat, skin impedance) captured by special hardware devices added to the sensor (i.e., hardware-based solutions that increase the cost of the sensors, and are more intrusive to the user). Unlike previously presented methods, the proposed technique classifies each image in terms of features extracted from just that image, and not from different samples of the fingerprint. This way the acquisition process is faster and more convenient to the final user (that does not need to keep his finger on the sensor for a few seconds, or place it several times).

Liveness detection solutions such as the one presented in this work are of great importance in the biometric field as they help to prevent direct attacks (those carried out with synthetic traits, and very difficult to detect), thus enhancing the level of security offered to the user.

## ACKNOWLEDGMENT

J. G. is supported by a FPU Fellowship from the Spanish MEC, F. A.-F. is supported by a Juan de la Cierva Fellowship from the Spanish MICINN, and J. F. is supported by a Marie Curie Fellowship from the European Commission. This work was supported by Spanish MEC under project TEC2006-13141-C03-03.

## REFERENCES

[1] T. Matsumoto, H. Matsumoto, K. Yamada, and H. Hoshino, "Impact of artificial gummy fingers on fingerprint systems," in *Proc. SPIE*, vol. 4677, 2002, pp. 275–289.
[2] V. Ruiz-Albacete, P. Tome-Gonzalez, F. Alonso-Fernandez, J. Galbally, J. Fierrez, and J. Ortega-Garcia, "Direct attacks using fake images in iris verification," in *in Proc. COST 2101 Workshop on Biometrics and Identity Management, BIOID*, 2008.
[3] M. Martinez-Diaz, J. Fierrez, F. Alonso-Fernandez, J. Ortega-Garcia, and J. A. Siguenza, "Hill-climbing and brute force attacks on biometric systems: a case study in match-on-card fingerprint verification," in *Proc. IEEE of International Carnahan Conference on Security Technology (ICCST)*, 2006, pp. 151–159.
[4] CC, "Common Criteria for Information Technology Security Evaluation. v3.1," 2006.
[5] BEM, "Biometric Evaluation Methodology. v1.0," 2002.
[6] T. Van der Putte and J. Keuning, "Biometrical fingerprint recognition: don't get your fingers burned," in *Proc. IFIP*, 2000, pp. 289–303.
[7] J. Galbally, R. Cappelli, A. Lumini, D. Maltoni, and J. Fierrez, "Fake fingertip generation from a minutiae template," in *Proc. ICPR*, 2008.
[8] J. Galbally, J. Fierrez, J. D. Rodriguez-Gonzalez, F. Alonso-Fernandez, J. Ortega-Garcia, and M. Tapiador, "On the vulnerability of fingerprint verification systems to fake fingerprint attacks," in *Proc. IEEE of International Carnahan Conference on Security Technology (ICCST)*, 2006, pp. 130–136.
[9] J. Galbally, J. Fierrez, and J. Ortega-Garcia, "Bayesian hill-climbing attack and its application to signature verification," in *Proc. IAPR International Conference on Biometrics (ICB)*. Springer LNCS-4642, 2007, pp. 386–395.
[10] B. Tan and S. Schuckers, "Comparison of ridge- and intensity-based perspiration liveness detection methods in fingerprint scanners," in *Proc. SPIE*, vol. 6202, 2006, p. 62020A.
[11] A. Antonelli, R. Capelli, D. Maio, and D. Maltoni, "Fake finger detection by skin distortion analysis," *IEEE Trans. on Information Forensics and Security*, vol. 1, pp. 360–373, 2006.
[12] Y. Chen and A. K. Jain, "Fingerprint deformation for spoof detection," in *Proc. IEEE BSYM*, 2005, pp. 19–21.
[13] P. Lapsley, J. Less, D. Pare, and N. Hoffman, "Anti-fraud biometric sensor that accurately detects blood flow," US Patent, 1998.
[14] D. Baldiserra, A. Franco, D. Maio, and D. Maltoni, "Fake fingerprint detection by odor analysis," in *Proc. IAPR ICB*. Springer LNCS-3832, 2006, pp. 265–272.
[15] L. Biel, O. Pettersson, L. Philipson, and P. Wide, "ECG analysis: A new approach in human identification," *IEEE Trans. on Instrumentation and Measurement*, vol. 50, pp. 808–812, 2001.
[16] P. Coli, G. L. Marcialis, and F. Roli, "Fingerprint silicon replicas: static and dynamic features for vitality detection using an optical capture device," *Int. Journal of Image and Graphics*, pp. 495–512, 2008.
[17] http://prag.diee.unica.it/LivDet09/.
[18] L. Shen, A. Kot, and W. Koo, "Quality measures of fingerprint images," *Proc. 3rd International Conference on Audio- and Video-Based Biometric Person Authentication, AVBPA*, vol. Springer LNCS-2091, pp. 266–271, 2001.
[19] F. Alonso-Fernandez, J. Fierrez, J. Ortega-Garcia, J. Gonzalez-Rodriguez, H. Fronthaler, K. Kollreider, and J. Bigun, "A comparative study of fingerprint image quality estimation methods," *IEEE Trans. on Information Forensics and Security*, vol. 2, no. 4, pp. 734–743, December 2007.
[20] J. Bigun, *Vision with Direction*. Springer, 2006.
[21] E. Lim, X. Jiang, and W. Yau, "Fingerprint quality and validity analysis," *Proc. International Conference on Image Processing, ICIP*, vol. 1, pp. 469–472, 2002.
[22] Y. Chen, S. Dass, and A. Jain, "Fingerprint quality indices for predicting authentication performance," *Proc. International Conference on Audio- and Video-Based Biometric Person Authentication, AVBPA*, vol. Springer LNCS-3546, pp. 160–170, 2005.
[23] T. Chen, X. Jiang, and W. Yau, "Fingerprint image quality analysis," *Proc. International Conference on Image Processing, ICIP*, vol. 2, pp. 1253–1256, 2004.
[24] L. Hong, Y. Wan, and A. Jain, "Fingerprint imagen enhancement: Algorithm and performance evaluation," *IEEE Trans. on Pattern Analysis and Machine Intelligence*, vol. 20, no. 8, pp. 777–789, August 1998.
[25] R. O. Duda, P. E. Hart, and D. G. Stork, *Pattern Classification*. Wiley, 2001.